\documentclass[11pt]{article}
\usepackage{acl2016}
\usepackage{times}
\usepackage{latexsym}

\usepackage[round]{natbib}

\usepackage{array}

\usepackage{times}
\usepackage{latexsym}

\usepackage{float}

\usepackage{amsmath}
\usepackage{amsfonts}
\usepackage{amssymb}

\usepackage{times}
\usepackage{url}
\usepackage{latexsym}
\usepackage{graphicx}
\usepackage{subfig}
\usepackage{amsmath}
\usepackage{amsfonts}
\usepackage{sidecap}
\usepackage{cancel}
\usepackage{enumitem}

\usepackage{pdfsync}
\usepackage{booktabs}
\usepackage{amssymb}
\usepackage{bm}
\usepackage{xcolor}

\newcommand{\ignore}[1]{}

\DeclareMathOperator*{\aggr}{Aggr}

\aclfinalcopy % Uncomment this line for the final submission
\title{Neural Network-based Word Alignment through Score Aggregation}

 \author{Jo\"el Legrand$^{1,2,\dag}$ \and Michael Auli$^{3}$ \and Ronan Collobert$^{3}$ \vspace{0.2cm}\\
          $^1$ Idiap Research Institute, Martigny, Switzerland \vspace{0.2cm}\\
         $^2$ Ecole Polytechnique F\'ed\'erale de Lausanne (EPFL), Lausanne, Switzerland \vspace{0.2cm}\\
         $^3$ Facebook AI Research, Menlo Park}

\date{}

\begin{document}

\maketitle

{\let\thefootnote\relax\footnotetext{$^{\dag}$This work was conducted while the first author did an internship at Facebook AI Research.}}

\begin{abstract}
We present a simple neural network for word alignment that builds source and
target word window representations to compute alignment scores for sentence pairs.
To enable unsupervised training, we use an aggregation operation that summarizes
the alignment scores for a given target word.
A soft-margin objective increases scores for true target words while
decreasing scores for target words that are not present.
Compared to the popular Fast Align model, our approach improves
alignment accuracy by 7 AER on English-Czech, by 6 AER on Romanian-English
and by 1.7 AER on English-French alignment.
%On English-French data we match  model and we outperform it on Romanian-English by 2.8 AER.
\end{abstract}

\section{Introduction}

Word alignment is the task of finding the correspondence between source and
target words in a pair of sentences that are translations of each other.
Generative models for this task
\citep{Brown:1990:SAM:92858.92860,Och:2003:SCV:778822.778824,Vogel1996COLING}
still form the basis for many machine translation systems
\citep{Koehn:2003:SPT:1073445.1073462,Chiang2007}.
% Although there have been several discriminative extensions
% \cite{Moore2005NAACL,Taskar2005NAACL,Blunsom2006ACL}.
% supervised data which is scarce even for high-resource language-pairs.

Recent neural approaches include \cite{yang-EtAl:2013:ACL2013} who introduce a feed-forward
network-based model trained on alignments that were generated by a
traditional generative model.
This treats potentially erroneous alignments as supervision.
\cite{tamura2014recurrent} sidesteps this issue by negative sampling
to train a recurrent-neural network on unlabeled data.
They optimize a global loss that requires an expensive beam search to
approximate the sum over all alignments.

In this paper we introduce a word alignment model that is simpler
in structure and which relies on a more tractable training procedure.
Our model is a neural network that extracts context information from
source and target sentences and then computes simple dot products to
estimate alignment links.
Our objective function is word-factored and does not require the
expensive computation associated with global loss functions.
The model can be easily trained on unlabeled data via a novel but simple
\emph{aggregation operation} which has been successfully applied
in the computer vision literature \citep{pinheiro:2015a}.
The aggregation combines the scores of all source words for a particular
target word and promotes
%, together with our soft-margin criterion,
source words which are likely to be aligned with a given target word
according to the knowledge the model has learned so far.
At test time, the aggregation operation is removed and source words
are aligned to target words by choosing the highest scoring candidates
(\textsection\ref{section:model}, \textsection\ref{section:architecture}).

We evaluate several forms for our aggregation operation such as computing the sum,
max and LogSumExp over alignment scores. Results on English-French,
English-Romanian, and Czech-English alignment show that our model
significantly  outperforms Fast Align, a popular log-linear reparameterization of IBM
Model 2 (Dyer et al., 2013; \textsection\ref{section:experiments})\nocite{Dyer13asimple}.

\section{Aggregation Model}
\label{section:model}

\begin{figure*}
\includegraphics[width=\textwidth]{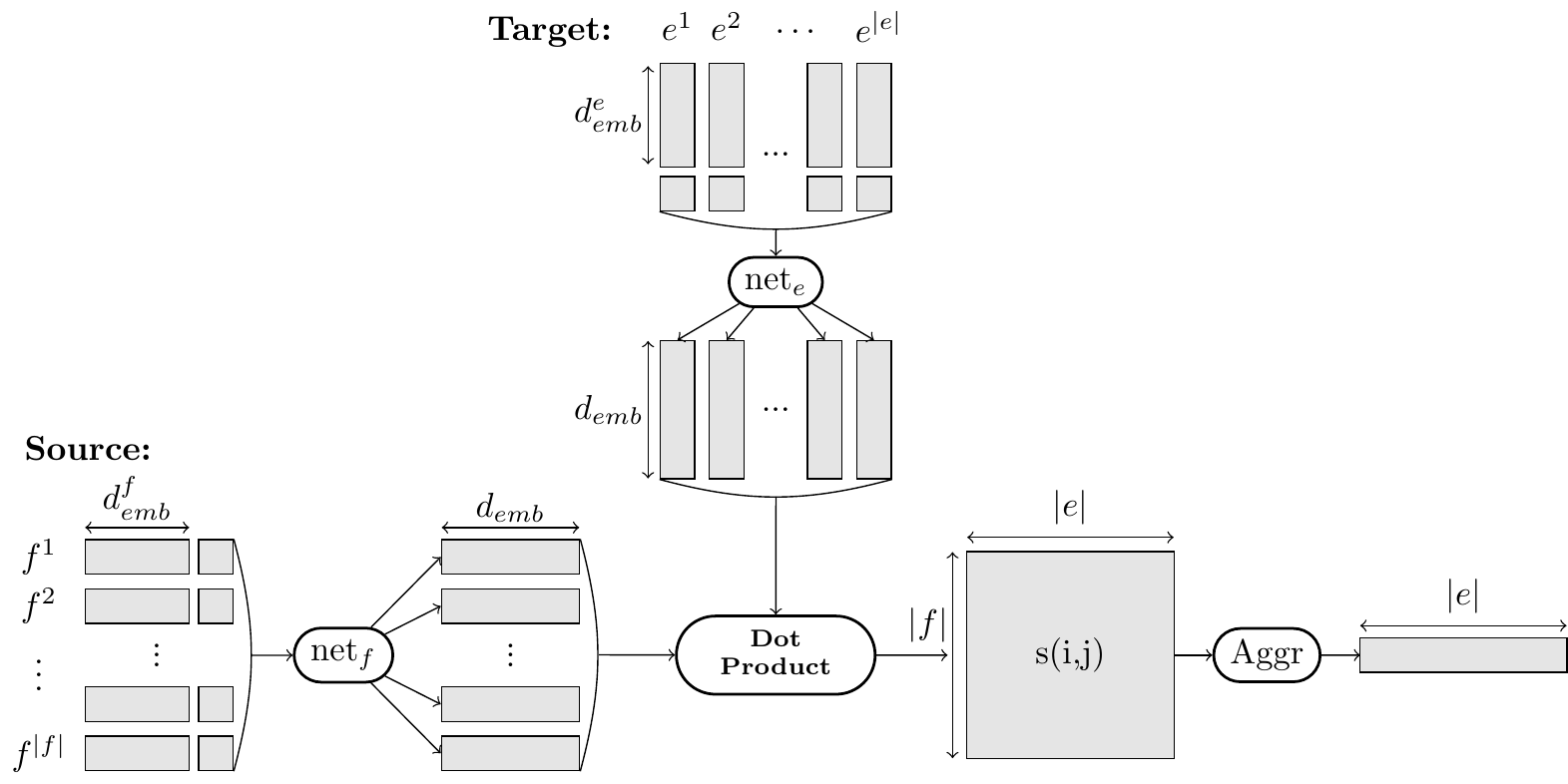}
\caption{\label{fig:model} Illustration of the model. The two networks $\textrm{net}_e$ and $\textrm{net}_f$ compute representations for source and target words. The score of an alignment link is a simple dot product between those source and target word representations. The aggregation operation summarizes the alignment scores for each target word.}
\end{figure*}

In the following, we consider a target-source sentence pair $(\mathbf{e},\,\mathbf{f})$, with $\mathbf{e} = (e_1,\,\dots,\, e_{|\mathbf{e}|})$ and $\mathbf{f} = (f_1,\,\dots,\, f_{|\mathbf{f}|})$. Words are represented by $f_j$ and $e_i$, which are indices in source and target dictionaries. For simplicity, we assume here that word indices are the only feature fed to our architecture. Given a source word $f_j$ and a target word $e_i$, our architecture embeds a window (of size $d^f_{win}$ and $d^e_{win}$, respectively) centered around each of these words into a $d_{emb}$-dimensional vector space. The embedding operation is performed with two distinct neural networks:
\begin{equation*}
\textrm{net}_e(\left[\mathbf{e}\right]_i^{d^e_{win}})\in \mathbb{R}^{d_{emb}} \end{equation*}
and
\begin{equation*}
\textrm{net}_f(\left[\mathbf{f}\right]_j^{d^f_{win}})\in \mathbb{R}^{d_{emb}}\,,
\end{equation*}

\noindent
where we denote the window operator as $$\left[\mathbf{x}\right]_i^d = (x_{i-d/2},\dots,x_{i+d/2})\,.$$
The matching score between a source word $f_j$ and a target word $e_i$ is then given by the dot-product:
\begin{equation}
\label{eq-matching-score}
s(i,\,j) = \textrm{net}_e(\left[\mathbf{e}\right]_i^{d^e_{win}}) \cdot \textrm{net}_f(\left[\mathbf{f}\right]_j^{d^f_{win}})\,.
\end{equation}
If $e_i$ is aligned to $f_{a_i}$, the score $s(i,\,a_i)$ should be high, while scores $s(i,\,j)\ \forall j\neq a_i$ should be low.

\subsection{Unsupervised Training}
\label{section:unsup}

In this paper, we consider an unsupervised setup where the alignment is not known at training time. We thus cannot minimize or maximize matching scores~(\ref{eq-matching-score}) in a direct manner. Instead, given a target word $e_i$ we consider the aggregated matching scores over the source sentence:
\begin{equation}
\label{eq-aggr}
s_{aggr}(i, \mathbf{f}) = \aggr_{j=1}^{|\mathbf{f}|} s(i,\,j)\,,
\end{equation}
where $\aggr$ is an aggregation operator (\textsection\ref{section:aggregation}). Consider a matching (positive) sentence pair $(\mathbf{e}^+, \mathbf{f})$ and a negative sentence pair $(\mathbf{e^-}, \mathbf{f})$. Given a word at index $i^+$ in the positive target sentence, we want to maximize the aggregated score $s_{aggr}(i^+, \mathbf{f})$ ($1 \leq i^+ \leq |\mathbf{e^+}|$) because we know it should be aligned to at least one source word.\footnote{We discuss how we handle unaligned target words in \textsection\ref{section:decoding}. Also, depending on the decoding algorithm the model can be used to predict many-to-many alignments.}
Conversely, given a word at index $i^-$ in the negative target sentence, we want to minimize $s_{aggr}(i^-, \mathbf{f})$ ($1 \leq i^- \leq |\mathbf{e^-}|$) because it is unlikely that the source sentence can explain the negative target word.
% it is unlikely that it will be aligned to any word in the source sentence:
Following these principles, we consider a simple soft-margin loss:
\begin{align}
{\cal L}(\mathbf{e}^+,\, \mathbf{e}^-,\, \mathbf{f}) & = \sum_{i^+=1}^{|\mathbf{e^+}|} \log(1 + e^{-s_{aggr}(i^+, \mathbf{f})})\nonumber\\
 & + \sum_{i^-=1}^{|\mathbf{e^-}|} \log(1 + e^{+s_{aggr}(i^-, \mathbf{f})})\,.\label{eq-loss}
\end{align}
Training is achieved by minimizing~(\ref{eq-loss}) and by sampling over triplets $(\mathbf{e}^+,\, \mathbf{e}^-,\, \mathbf{f})$ from the training data.

\subsection{Choosing the Aggregation}
\label{section:aggregation}

The aggregation operation~(\ref{eq-aggr}) is only present during training and acts as a filter which aims to explain a given target word $e_i$ by one or more source words. If we had the word alignments, then we would sum over the source words $f_j$ aligned with $e_i$. However, in our setup alignments are not available at training time, so we must rely on what the model has learned so far to filter the source words.
%\mamark{, similar to the classical EM algorithm \cite{dempster2007}}.
We consider the following strategies:
\begin{itemize}
\item \textbf{Sum:} ignore the knowledge learned so far, and assign the same weight to all source words $f_j$ to explain $e_i$.\footnote{This can be seen by observing that the gradients for all source words are the same.} In this case, we have
$$
s_{aggr}(i, \mathbf{f}) = \sum_{j=1}^{|\textbf{f}|} s(i,\,j)\,.
$$
\item \textbf{Max:} encourage the best aligned source word $f_j$, according to what the model has learned so far. In this case, the aggregation is written as:
$$
s_{aggr}(i, \mathbf{f}) = \max_{j=1}^{|\textbf{f}|} s(i,\,j)\,.
$$
\item \textbf{LSE:} give similar weights to source words with similar scores. This can be achieved with a LogSumExp aggregation operation (also called LogAdd), and is defined as:
%\cite{Boyd:2004:CO:993483} LSE is much older than this
\begin{equation}
\label{eq-lse}
s_{aggr}(i, \mathbf{f}) = \frac{1}{r} \log\left( \sum_{j=1}^{|\textbf{f}|} e^{r\,s(i,\,j)} \right)\,,
\end{equation}
where $r$ is a positive scalar (to be chosen) controlling the smoothness of the aggregation. For small $r$, the aggregation is equivalent to a sum, and for large $r$, the aggregation acts as a max.
\end{itemize}

\subsection{Decoding}
\label{section:decoding}

At test time, we align each target word $e_i$ with the source word $f_j$ for which the matching score $s(i,\,j)$ in~(\ref{eq-matching-score}) is highest.\footnote{This may result in a source word being aligned to multiple target words.} However, not every target word is aligned, so we consider only alignments with a matching score above a threshold:
 \begin{equation}
 \label{eq-thresholding}
 s(i,\,j) > \mu^{-}(e_i) + \alpha \, \sigma^-(e_i)\,,
 \end{equation}
 where $\alpha$ is a tunable hyper-parameter, and $$\mu^{-}(e_i) = \mathop{{}\mathbb{E}}_{\{\tilde{e}_k=e_i \,\in\, \mathbf{\tilde{e}},\,\tilde{f}_{j^-}\, \in\, \mathbf{\tilde{f}-}\}} \left[ s(k,\, j^-) \right]$$ is the expectation over all training sentences $\tilde{\mathbf{e}}$ containing the word $e_i$, and all words $\tilde{f}^-_j$ belonging to a corresponding negative source sentence $\mathbf{\tilde{f}}^-$, and $\sigma^-(e_i)$ is the respective variance.

\section{Neural Network Architecture}
\label{section:architecture}

Our model consists of two convolutional neural networks $\textrm{net}_e$ and $\textrm{net}_f$ as shown in~(\ref{eq-matching-score}). Both of them take the same form, so we detail only the target architecture.

\subsection{Word embeddings}

The discrete features $\left[\mathbf{e}\right]_i^{d^e_{win}}$ are embedded into a $d^e_{emb}$-dimensional vector space via a lookup-table operation as first introduced
in \cite{Bengio2000NIPS}:
\begin{align*}
x_i^e & = \textsc{LT}_{W^e}(\left[\mathbf{e}\right]_i^{d^e_{win}}) \\
    & = (\textsc{LT}_{W^e}(e_{i-d^e_{win}/2}),\,\dots,\, \textsc{LT}_{W^e}(e_{i+d^e_{win}/2}))\,,
\end{align*}
where the lookup-table operation applied at index $k$ returns the $k^{th}$ column of the parameter matrix $W^e$:
$$
\textsc{LT}_{W^e}(k) = W^e_{\bullet,\,k}\,.
$$
The matrix $W^e$ is of size $|{\cal V}^e| \times d^e_{emb}$, where ${\cal V}^e$ is the target vocabulary, and $d_{emb}^e$ is the word embedding size for the target words.

\subsection{Convolutional layers}

The word embeddings output by the lookup-table are concatenated and fed through two successive 1-D convolution layers.
The convolutions use a step size of one and extract context features for each word.
The kernel sizes $k_1^e$ and $k_2^e$ determine the size of the window $d^e_{win} = k_1^e + k_2^e - 1$ over which features will be extracted by $\textrm{net}_e$.
In order to obtain windows centered around each word, we add $(k_1^e + k_2^e) / 2 - 1$ padding words at the beginning and at the end of each sentence.

The first layer $cnn^e$ applies the linear transformation $M^{e,1}$ exactly $k_2^e$ times to consecutive spans of size $k_1^e$ to the $d^e_{win}$ words in a given window:
% The first layer $cnn^e$ pads the input with zero values to the left and right and applies a 1-D convolution over $x^e$ with kernel width $k_1^e$:
%
% \begin{flalign*}
% & cnn^e(x^e) =\\
% & \:\:\:\:\:\:\:\:\:M^{e,1}\begin{pmatrix}
%   \Big\{e_{i-\frac{d^e_{win}}{2}}\Big\}^e, \dots, \Big\{e_{i-\frac{d^e_{win}}{2}+k_1^e}\Big\}^e \\
%   \large{\vdots} \\
%   \Big\{e_{i-\frac{d^e_{win}}{2}-k_1^e}\Big\}^e, \dots, \Big\{e_{i-\frac{d^e_{win}}{2}}\Big\}^e\end{pmatrix},
% \end{flalign*}

\begin{flalign*}
& cnn^e(x_i^e) = M^{e,1}\begin{pmatrix}
  \textsc{LT}_{W^e}(\left[\mathbf{e}\right]^{k_1^e}_{i-a}) \\
  \large{\vdots} \\
  \textsc{LT}_{W^e}(\left[\mathbf{e}\right]^{k_1^e}_{i+a}) \\
%   x^e_{i-w}, \dots, x^e_{i-w+k_1^e} \\
%   \large{\vdots} \\
%   x_{i-w-k_1^e}^e, \dots, x_{i-w^e}
\end{pmatrix},
\end{flalign*}

where $a = \lfloor\frac{k^e_2}{2}\rfloor$, $M^{e,1}\in \mathbb{R}^{d^e_{hu} \times (d^e_{emb}\,k^e_1)}$ is a matrix of parameters, and $d^e_{hu}$ is the number of hidden units ($hu$). The outputs of the first layer $cnn^e$ are concatenated to form a matrix of size $k^e_2\textrm{ }d^e_{hu}$ which is fed to the second layer:

\begin{equation}
\label{eq-neural-net}
\textrm{net}_e(x_i^e) = M^{e,2} \, \tanh(cnn^e(x_i^e))\,
\end{equation}

% \begin{equation}
% \label{eq-neural-net}
% \textrm{net}_e(x^e) = M^{e,2} \, \tanh(M^{e,1}\, x^e)\,
% \end{equation}

where $M^{e,2} \in \mathbb{R}^{d_{emb} \times (k^e_2~d^e_{hu})}$ is a matrix of parameters, and the $tanh(\cdot)$ operation is applied element wise.
The parameters $W^e$, $M^{e,1}$ and $M^{e,2}$ are trained by stochastic gradient descent to minimize the loss~(\ref{eq-loss}) introduced in \textsection\ref{section:unsup}.

\subsection{Additional Features}
\label{section:features}

In addition to the raw word indices, we consider two additional discrete features which were handled in the same way as word features by introducing an additional lookup-table for each of them. The output of all lookup-tables was concatenated, and fed to the two-layer neural network architecture~(\ref{eq-neural-net}).
\begin{description}[leftmargin=0pt]

\item[Distance to the diagonal.] This feature can be computed for a target word $e_i$ and a source word $f_j$:
$$diag(i, j) = \left| \frac{i}{|\mathbf{e}|} - \frac{j}{|\mathbf{f}|}\right|\,,$$
This feature allows the model to learn that aligned sentence pairs use roughly the same word order and that alignment links remain close to the diagonal. We use this feature only for the source network because it encodes relative position information which only needs to be encoded once. If we would use absolute position instead, then we would need to encode this information both on the source and the target side.

\item[Part-of-speech] Words pairs that are good translations of each other are likely to carry the same part of speech in both languages \citep{W95-0115}. We therefore add the part-of-speech information to the model.

\item[Char n-gram.] We consider unigram character position features. Let $K$ be the maximum size for a word in a dictionary. We denote the dictionary of characters as $\mathcal{C}$. Every character is represented by its index $c$ (with $1<c<|\mathcal{C}|$). We associate every character $c$ at position $k$ with a vector at position $((k-1)*|\mathcal{C}|) + c$ in a lookup-table. For a given word, we extract all unigram character position embeddings, and average them to obtain a character embedding for a given word.

%We consider unigram, bigram and trigram character features for which we have separate lookup-tables; we also use a special character marking word boundaries. For a given word, we extract all character n-grams of order $n$, then lookup their embeddings, and average them. This results in three averaged character embedding representations per word that are input to the source and target networks.
\end{description}

\section{Experiments}
\label{section:experiments}

\subsection{Datasets}

We use the English-French Hansards corpus as distributed by
the NAACL 2003 shared task \citep{mihalcea-pedersen:2003:Partext}.
This dataset contains 1.1M sentence pairs and the test and validation sets contain
447 and 37 examples respectively. We also evaluate on the Romanian-English dataset of the ACL 2005 shared task \citep{martin2005acl} comprising 48K sentence pairs for training, 248 for testing and 17 for validation. For English-Czech experiments,
we use the WMT news commentary corpus for training (150K sentence pairs)
and a set of 515 sentences for testing \citep{bojar:prokopova:2006}.

\subsection{Evaluation}

Our models are evaluated in terms of precision, recall, F-measure and Alignment Error Rate (AER).
We train models in each language direction and then symmetrize the resulting alignments using either the \emph{intersection} or the
\emph{grow-diag-final-and} heuristic \citep{Och:2003:SCV:778822.778824,Koehn:2003:SPT:1073445.1073462}.
We validated the choice of symmetrization heuristic on each language pair and chose the best one for each model considering the two aforementioned types as well as \emph{grow-diag-final} and \emph{grow-diag}.

Additionally, we train phrase-based machine translation models with our alignments using the popular Moses toolkit \citep{Koehn:2007:MOS:1557769.1557821}. For English-French, we train on the news commentary corpus v10, for English-Czech we used news commentary corpus v11, and for Romanian-English we used the Europarl corpus v8. We tuned our models on the WMT2015 test set for English-Czech as well as for Romanian-English; for English-French we tuned on the WMT2014 test set. Final results are reported on the WMT2016 test set for English-Czech as well as Romanian-English, and for English-French we report results on the WMT2015 test set (as there is no track for this language-pair in 2016).

We compare our model to Fast Align, a popular log-linear reparameterization of IBM Model 2 \citep{Dyer13asimple}.

\subsection{Setup}
\label{setup}
The kernel sizes of the target network $\textrm{net}_e(\cdot)$ are set to $k^e_1 = k^e_2 = 3$ for all language pairs. The kernel sizes of the source network $\textrm{net}_f(\cdot)$
are set to $k^f_1 = k^f_2 = 3$ for Romanian-English as well as English-Czech; and for English-French we used $k^f_1 = k^f_2 = 1$.

The number of hidden units are $d^e_{hu} = d^f_{hu} = 256$ and $d_{emb}$ is set to 256, The source ${\cal V}_f$ and target ${\cal V}_e$ dictionaries consist of the 30K most common words for English, French and Romanian, and 80K for Czech. All other words are mapped to a unique \textit{UNK} token. The word embedding sizes $d^e_{emb}$ and $d^f_{emb}$, as well as the char-n-gram embedding size is $128$. For LSE, we set $r=1$ in~(\ref{eq-lse}).

We initialize the word embeddings with a simple PCA computed over the matrix of word co-occurrence counts \citep{Lebret14}. The co-occurrence counts were computed over the common crawl corpus provided by WMT16.
For part of speech tagging we used the Stanford parser on English-French data, and MarMoT \citep{mueller-schmid-schutze:2013:EMNLP} for Romanian-English as well as English-Czech.
%The Wikipedia corpora for English, French and Romanian comprise $1747M$, $444M$ and $61M$ tokens, respectively.

We trained 4 systems for the ensembles, each using a different random seed to vary the weight initialization as well as the shuffling of the training set. We averaged the alignment scores predicted by each system before decoding.
The alignment threshold variables $\mu^-(e_i)$ and $\sigma^-(e_i)$ for decoding (\textsection\ref{section:decoding}) were estimated on 1000 random training sentences, using 100 negative sentences for each of them. Words not appearing in this training subset were assigned $\mu^-(e_i)=\sigma^-(e_i)=0$.

For systems where $d^e_{win}>1$ and $d^f_{win}>1$, we saw a tendency of aligning frequent words regardless on if they appeared in the center of the context window or not. For instance, a common mistake would be to align "the \textit{cat} sat", with "PADDING \textit{le} chat".
To prevent such behavior, we occasionally replaced the center word in a target window by a random word during training. We do this for every second training example on average and we tuned this rate on the validation set.

%\mamark{How often did you replace the center word?}

\subsection{Results}

We first explore different choices for the aggregation operator (\textsection\ref{section:aggregation}), followed by an ablation to investigate the impact of the different additional features (\textsection\ref{section:features}). Next we compare to the Fast Align baseline. Finally, we evaluate our alignments within a full translation system for all language pairs.

\subsubsection{Aggregation operation}

Table~\ref{tab:aggr} shows that the LogSumExp (LSE) aggregator performs best on
all datasets for every direction as well as in the symmetrized setting using the grow-diag-final heuristic.
All results are based on a single model trained with the 'distance to the diagonal' feature detailed above.\footnote{We use kernel sizes $k^e_1 = k^e_2 = 3$ and $k^f_1 = k^f_2 = 1$ for all language pairs in this experiment.}
%Learning with the sum is most challenging on English-French whereas on Romanian-English both the sum and the max perform equally low.
We therefore use LSE for the remaining experiments.

\begin{table}[h]
\center
\begin{tabular}[tbph]{lrrr}
 & Max & Sum & LSE \\
\hline
En-Fr       & 18.1 & 23.0 & \textbf{15.1}\\
Fr-En       & 20.7 & 26.9 & \textbf{15.8}\\
symmetrized & 14.8 & 24.1 & \textbf{12.8}\\
\hline
Ro-En       & 42.2 & 42.0 & \textbf{37.6}\\
En-Ro       & 40.4 & 40.2 & \textbf{35.7}\\
symmetrized & 36.4 & 35.6 & \textbf{32.2}\\
\hline
En-Cz       & 27.9 & 35.6 & \textbf{24.5}\\
Cz-En       & 26.5 & 33.6 & \textbf{24.5}\\
symmetrized & 21.8 & 32.7 & \textbf{21.0}
\end{tabular}
\caption{\label{tab:aggr} Alignment error rates for different aggregation operations
in each language direction and with \emph{grow-diag-final-and} symmetrization.}
\end{table}

\subsubsection{Additional features}

Table \ref{tab:feat} shows the effect of the different input features. Both POS and the distance to the diagonal feature significantly improve accuracy. Position information via the 'distance to the diagonal' feature is helpful for all language pairs, and POS information is more effective for Romanian-English and English-Czech which involve morphologically rich languages.
We use the POS and 'distance to the diagonal feature' for the remaining experiments.

\begin{table*}
\center
\begin{tabular}{l|c|c|c||c|c|c||c|c|c|}
              & \multicolumn{3}{c||}{English-French} & \multicolumn{3}{c||}{Romanian-English} & \multicolumn{3}{c|}{English-Czech}\\
\cline{2-10}
              & En-Fr & Fr-En & sym  & Ro-En & En-Ro & sym  & En-Cz & Cz-En & sym  \\
\hline
words         & 22.2  & 24.2  & 15.7 & 47.0  & 45.5  & 40.3 & 36.9  & 36.3  & 29.5 \\
+ POS         & 20.9  & 23.9  & 15.3 & 45.3  & 42.9  & 36.9 & 35.6  & 33.7  & 28.2 \\
+ diag        & 15.1  & 15.8  & 12.8 & 37.6  & 35.7  & 32.2 & 24.8  & 24.5  & 21.0 \\
+ POS + diag  & \textbf{13.2}  & \textbf{12.1}  & \textbf{10.2} & \textbf{33.1} & \textbf{32.2} & \textbf{27.8} & \textbf{24.6} & \textbf{22.9} & \textbf{19.9} \\
\end{tabular}
\caption{\label{tab:feat} Alignment error rates using different input features
in each language direction and with \emph{grow-diag-final-and} symmetrization.}
\end{table*}

\subsubsection{Comparison with the baseline}

In the following results we label our model as NNSA (Neural network score aggregation).
On English-French data (Table~\ref{tab:french}) our model outperforms
the baseline \citep{Dyer13asimple} in each individual language direction
as well as for the symmetrized setting.
With an ensemble of four models, we outperform the baseline by 1.7 AER (from 11.4 to 9.7),
and with an individual model we outperform it by 1.2 AER (from 11.4 to 10.2).
Note that the choice of symmetrization heuristic greatly affects accuracy, both for the baseline
and NNSA.

\begin{table}
\center
\begin{tabular}[tbph]{lrrrr}
& P & R & F1 & AER \\
\hline
English-French & & & & \\
\hspace{0.05in} Baseline &  49.6 & 89.8 & 63.9 & 16.7 \\ %43.7 & 85.8 & 57.9 & 18.8 &
%Giza++ En-Fr & 51.1 & 94.3 & 66.3 & 11.4 \\
\hspace{0.05in} NNSA &  64.7 & 80.7 & 71.8 & 13.2\\ %58.7 & 81.4 & 68.2 & 13.7 &
%\textbf{This paper}$^{+}$ $\rightarrow$ & 60.6 & 76.9 & 67.8 & 17.2 \\
\hspace{0.05in} + ensemble & 61.5 & 85.8 & 71.6 & \textbf{11.6}\\
%\textbf{This paper}$^{+v}$ $\rightarrow$ & 62.4 & 76.7 & 68.8 & 16.7 \\
\hline
French-English & & & & \\
\hspace{0.05in} Baseline &  52.9 & 88.4 & 66.2 & 16.2\\ %45.0 & 84.9 & 58.9 & 17.3 &
%Giza++ Fr-En & 56.1 & 93.9 & 70.2 & 10.3 \\
\hspace{0.05in} NNSA &  61.7 & 86.3 & 72.0 & 12.1\\ %58.5 & 75.5 & 65.9 & 17.0 &
%\textbf{This paper}$^{+}$ $\leftarrow$ & 65.5 & 80.2 & 72.1 & 15.6\\
\hspace{0.05in} + ensemble  & 62.6 & 86.7 & 72.7 & \textbf{11.6}\\
%\textbf{This paper}$^{+v}$ $\leftarrow$ &  65.2 & 82.0 & 72.6 & \textbf{14.1}\\ %58.5 & 75.5 & 65.9 & 17.0 &
\hline
symmetrized & & & & \\
\hspace{0.05in} Baseline (inter) &  69.6 & 84.0 & 76.1 & 11.4 \\ %61.6 & 78.7 & 69.1 & 13.2 &
%\hspace{0.05in} Baseline (gdfa) &  47.7 & 92.6 & 62.9 & 15.3 \\ %61.6 & 78.7 & 69.1 & 13.2 &
%Giza++ (intersect) & 71.1 & 90.3 & 79.5 & 6.1 \\
%Giza++ (gdfa) & 50.7 & 96.3 & 66.5 & 9.9 \\
%\hspace{0.05in} NNSA (inter) & 76.5 & 70.0 & 73.1 & 16.8\\ %56.3 & 83.7 & 67.3 & 12.0
%\textbf{This paper}$^{+}$ $\leftrightarrow$ (inter) & 75.9 & 64.2 & 69.5 & 21.2 \\ %56.3 & 83.7 & 67.3 & 12.0 &
%\textbf{This paper}$^{+v}$ $\leftrightarrow$ (inter) & 76.4 & 62.1 & 68.5 & 22.3 \\ %56.3 & 83.7 & 67.3 &12.0 &
%\textbf{This paper}$^{v/+v}$ $\leftrightarrow$ (inter) & 76.5 & 66.4 & 71.1 & 19.7\\ %56.3 & 83.7 & 67.3
\hspace{0.05in} NNSA (gdfa) &  60.4 & 88.5 & 71.8 & 10.2\\ %56.3 & 83.7 & 67.3 & 12.0 &
%\textbf{This paper}$^{+}$ $\leftrightarrow$ (gdfa) & 55.1 & 84.7 & 66.8 & 15.2 \\ %56.3 & 83.7 & 67.3 & 12.0 &
% \textbf{This paper}$^{+v}$ $\leftrightarrow$ (gdfa) & 61.2 & 83.5 & 70.6 & 13.4\\ %56.3 & 83.7 & 67.3 &12.0 &
\hspace{0.05in} + ensemble & 59.3 & 89.9 & 71.4 & \textbf{9.7}\\ %56.3 & 83.7 & 67.3 &12.0 &
\end{tabular}
\caption{\label{tab:french} English-French results on the test set in terms of precision (P),
recall (R), F-score (F1) and AER; ensemble denotes a combination of four systems and
we use the \emph{intersection} (inter) and \emph{grow-diag-final-and} symmetrization (gdfa) heuristics.}
\end{table}

On Romanian-English (Table~\ref{tab:romanian}) our model outperforms
the baseline in both directions as well. Adding ensembles further improves accuracy and
leads to a significant improvement of 6 AER over the best symmetrized baseline result (from 32 to 26).

\begin{table}
\center
\begin{tabular}[tbph]{lrrrr}
& P & R & F1 & AER \\
\hline
Romanian-English & & & & \\
\hspace{0.05in} Baseline &  70.0 & 61.0 & 65.2 & 34.8 \\ %80.4 & 66.2 & 72.6 & 27.4 &
%Giza++ Ro-En & 74.1 & 65.0 & 69.3 & 30.7\\
%Ro-En &  75.8 & 59.9 & 66.9 & 33.1 \\ %old (with no target context)
\hspace{0.05in} NNSA & 75.1 & 65.2 & 69.8 & 30.2 \\ %new with target context
\hspace{0.05in} + ensemble & 75.8 & 62.8 & 68.7 & \textbf{31.3}\\
\hline
English-Romanian & & & & \\
\hspace{0.05in} Baseline &  71.3 & 60.8 & 65.6 & 34.4\\ % 77.9 & 60.4 & 68.0 & 32.0 &
%Giza++ En-Ro & 72.7 & 62.3 & 67.1 & 32.9\\
%En-Ro &  78.3 & 59.8 & 67.8 & 32.2\\%old (with no target context)
\hspace{0.05in} NNSA & 78.1 & 61.7 & 69.0 & 31.1\\%new with target context
\hspace{0.05in} + ensemble & 78.4 & 63.2 & 70.0 & \textbf{30.0}\\
\hline
%sym
symmetrized & & & & \\
%\hspace{0.05in} Baseline (inter) &  87.8 & 53.1 & 66.2 & 33.8\\ %79.0 & 70.1 & 74.3 & 25.7 &
\hspace{0.05in} Baseline (gdfa) &  69.5 & 66.5 & 68.0 & 32.0\\ %79.0 & 70.1 & 74.3 & 25.7 &
%Giza++ (intersect) & 93.6 & 52.5 & 67.2 & 32.8 \\
%Giza++ (gdfa) & 74.7 & 71.2 & 7.9 & 27.1 \\
%intersect & 92.3 & 46.5 & 61.9 & 38.1\\ %old (with no target )
%gdfa &  78.3 & 67.1 & 72.2 & 27.8\\%old (with target context)
%\hspace{0.05in} NNSA (inter) & 92.5 & 49.0 & 64.1 & 35.9\\ %old (with no target )
\hspace{0.05in} NNSA (gdfa) &  74.1 & 71.8 & 73.0 & 27.0\\%old (with target context)
%+ ensemble & 77.8 & 70.5 & 74.0 & \textbf{26.0} %old (with no target )
\hspace{0.05in} + ensemble & 73.0 & 74.5 & 73.7 & \textbf{26.0} \\
\end{tabular}
\caption{Romanian-English results (cf.~Table~\ref{tab:french}).}
\label{tab:romanian}
\end{table}

On English-Czech (Table~\ref{tab:czech}) our model outperforms the baseline in both directions as well.
We added the character feature to better deal with the morphologically rich nature of Czech and the feature reduced AER by 2.1 in the symmetrized setting.
An ensemble improved accuracy further and led to a 7 AER improvement over the best symmetrized baseline result (from 22.8 to 15.8).

\begin{table}
\center
\begin{tabular}[tbph]{lrrrr}
& P & R & F1 & AER \\
\hline
English-Czech & & & & \\
\hspace{0.05in} Baseline& 68.4 & 73.3 & 70.7 & 26.6 \\
%Giza++ En-Cz & 60.3 & 65.8 & 62.9 & 35.5\\
\hspace{0.05in} NNSA & 72.0 & 74.3 & 73.1 & 24.6\\
\hspace{0.05in} + char n-gram & 73.8 & 75.4 & 74.6 & 23.2\\
\hspace{0.05in} + ensemble & 78.8 & 77.2 & 78.0 & \textbf{20.0} \\
\hline
Czech-English & & & & \\
\hspace{0.05in} Baseline & 68.6 & 74.0 & 71.2 & 25.7 \\
%Giza++ Cz-En & 66.0 & 74.1 & 69.8 & 26.7 \\
\hspace{0.05in} NNSA & 74.1 & 74.0 & 74.0 & 22.9\\
\hspace{0.05in} + char n-gram & 78.1 & 74.1 & 76.1 & 21.4\\
\hspace{0.05in} + ensemble & 79.1 & 77.7 & 78.4 & \textbf{18.7} \\
\hline
symmetrized & & & & \\
\hspace{0.05in} Baseline (inter) & 88.1 & 66.6 & 76.0 & 22.8\\
%\hspace{0.05in} Baseline (gdfa) & 65.9 & 78.6 & 71.7 & 24.5 \\
%Giza++ (intersect) & 94.6 & 56.3 & 70.6 & 28.8\\
%Giza++ (gdfa) & 65.8 & 78.1 & 71.4 & 24.9\\
%\hspace{0.05in} NNSA (inter) & 93.1 & 59.9 & 72.9 & 26.3\\
\hspace{0.05in} NNSA (gdfa) & 75.7 & 80.3 & 76.3 & 19.9 \\
\hspace{0.05in} + char n-gram & 76.9 & 81.3 & 79.1 & 17.8 \\
\hspace{0.05in} + ensemble & 78.9 & 83.2 & 81.0 & \textbf{15.8}\\
\end{tabular}
\caption{\label{tab:czech} Czech-English results (cf.~Table~\ref{tab:french}).}
\end{table}

\subsubsection{BLEU evaluation}

Table \ref{tab:bleu} presents the BLEU evaluation of our alignments. For each language-pair, we select the best alignment model reported in Tables \ref{tab:french}, \ref{tab:romanian} and \ref{tab:czech}, and align the training data. We use the alignments to run the standard phrase-based training pipeline using those alignments.
Our BLEU results show the average BLEU score and standard deviation for five runs of minimum error rate training (MERT; Och 2003)\nocite{Och2003MinimumER}.

Our alignments achieve slightly better results for Romanian-English as well as English-Czech while performing on par with Fast Align on English-French translation.

% \begin{table}[h]
% \center
% \begin{tabular}[tbph]{l|c|c}
% & Baseline & Our model \\
% \hline
% Fr-En & $25.4 \pm 0.1$ & $\mathbf{25.5} \pm 0.1$\\
% Ro-En & $ 22.4 \pm 0.0 $ & $\mathbf{22.5} \pm 0.0$\\
% Cs-En & $16.9 \pm 0.1$ & $\mathbf{17.2} \pm 0.1$\\

% \end{tabular}
% \caption{Bleu score obtained using Moses (with the default parameters) for the different alignment models and pairs of languages}
% \label{tab:bleu}
% \end{table}

\begin{table}[h]
\center
\begin{tabular}[tbph]{l|c|c}
& Baseline & NNSA \\
\hline
French-English & $25.4 \pm 0.1$ & $\mathbf{25.5} \pm 0.1$\\
Romanian-English & $ 21.3 \pm 0.1 $ & $\mathbf{21.6} \pm 0.1$\\
Czech-English & $17.2 \pm 0.1$ & $\mathbf{17.6} \pm 0.1$\\

\end{tabular}
\caption{Average BLEU score and standard deviation for five runs of MERT.}
\label{tab:bleu}
\end{table}

\section{Analysis}

In this section, we analyze the word representations learned by our model. We first focus on the source representations: given a source window, we obtain its distributional representation and then compute the Euclidean distance to all other source windows in the training corpus. Table \ref{tab:analysis2} shows the nearest windows for two source windows; the closest windows tend to have similar meanings.

We then analyze the relation between source and target representations: given a source window we compute the alignment scores for all target sentences in the training corpus.
Table \ref{tab:analysis1} shows for two source windows which target words have the largest alignment scores.
The example "in working together" is particularly interesting since the aligned target words \emph{collabore}, \emph{coordon\'es}, and \emph{concert\'es} mean \emph{collaborate}, \emph{coordinated}, and \emph{concerted}, which all carry the same meaning as the source window phrase.

\begin{table}[h]
\center
\begin{tabular}{c|c}
 \multicolumn{1}{c|}{the voting process} & \multicolumn{1}{c}{in working together}    \\
 % & \multicolumn{1}{c|}{canada child tax}       & future economic opportunities \\
\hline
the voting area         & for working together    \\%& the child tax & new economic opportunities\\
the voting power        & with working together   \\%& canadian child tax& the economic opportunities\\
the voting rules        & from working together   \\%& natinal child benefit   & important economic opportunities\\
the voting system       & about working together  \\%& federal child tax & real economic incentives\\
the voting patterns     & by working together     \\%& national child tax & many economic opportunities\\
%the voting arrangements & to working together     \\%& new child tax & basic economic needs\\
the voting ballots      & and working together    \\%& existing child tax & specific economic policies\\
their voting patterns   & while working together  \\%& canadian child benefit & the economic chalenges\\
%our voting procedures   & us working together     \\%& federal child benefit & the economic consequences\\
%, votins practices      & canada working together \\%& million child tax &\\
\end{tabular}
\caption{\label{tab:analysis2} Analysis of source window representations. Each column shows a window over the source sentence followed by several close neighbors in terms of Euclidean distance (among the 30 nearest).}
\end{table}

\begin{table}[h]
\center
\begin{tabular}{c|c}
the voting process & in working together  \\%& canada child tax & future economic opportunities\\
\hline
vote               & travaill\'e          \\%& enfant & \'economiques\\
voteraient         & travailleront        \\%& enfance          && \'economie\\
votent             & collaboration        \\%&& infantil         & \'economiques\\
voter              & travaillant          \\%& maternelle       && perspectives\\
votant             & oeuvrant             \\%& enfant           && d\'eveloppements\\
scrutin            & concertés            \\%& prestation       && r\'eam\'enagements\\
suffrage           & coordon\'es            \\%& descendance      && expansion\\
proc\'edure        & concert              \\%& subvention       && plans\\
investiture        & collabore            \\%& canada           && previsions\\
\'elections        & coop\'eration          \\%& loyer            && croissance\\
\end{tabular}
\caption{\label{tab:analysis1} Analysis of source and target representations. Each column shows a source window and the target words which are most aligned according to our model.}
\end{table}

\section{Conclusion}

In this paper, we present a simple neural network
alignment model trained on unlabeled data.
Our model computes alignment scores as dot products between
representations of windows around source and target words.
We apply an \emph{aggregation operation} borrowed from the computer
vision literature to make unsupervised training possible.
The aggregation operation acts as a filter over alignment scores
and allows us to determine which source words explain a given
target word.

We improve over Fast Align, a popular log-linear reparameterization of IBM Model 2
\citep{Dyer13asimple} by up to 6 AER on Romanian-English, 7 AER on English-Czech data and 1.7 AER on English-French alignment. Furthermore, we evaluated our model as part of a full machine translation pipeline and showed that our alignments are better or on par compared to Fast Align in terms of BLEU.

%show  while using in a full translation pipeline.

\bibliography{paper}
\bibliographystyle{plainnat}
%\bibliography{acl2016}
%\bibliographystyle{acl2016}

\appendix

\end{document}